%% file: main.tex
\documentclass[letterpaper, 10 pt, conference]{ieeeconf}  

\IEEEoverridecommandlockouts                              

\newcounter{myromancnt}

\input{format}

\UseRawInputEncoding
\title{\LARGE \bf
Actuator-Constrained Reinforcement Learning for High-Speed Quadrupedal Locomotion}
\author{Young-Ha Shin$^{1}$, Tae-Gyu Song$^{1}$, Gwanghyeon Ji$^{1}$, and Hae-Won Park$^{1}$, \textit{Member, IEEE}
\thanks{This research was supported by the Challengeable Future Defense Technology Research and Development Program through the Agency For Defense Development(ADD) funded by the Defense Acquisition Program Administration(DAPA) in 2022(No.915027201)}
\thanks{$^{1}$Young-Ha Shin, Tae-Gyu Song, Gwanghyeon Ji and Hae-Won Park are with the Humanoid Robot Research Center, Department of Mechanical Engineering, Korea Advanced Institute of Science and Technology, Yuseong-gu, Daejeon 34141, Republic of Korea. {\tt\small haewonpark@kaist.ac.kr}}
}

\begin{document}
\maketitle
\thispagestyle{empty}
\pagestyle{empty}

\begin{abstract}
This paper presents a method for achieving high-speed running of a quadruped robot by considering the actuator torque-speed operating region in reinforcement learning. The physical properties and constraints of the actuator are included in the training process to reduce state transitions that are infeasible in the real world due to  motor torque-speed limitations. The gait reward is designed to distribute motor torque evenly across all legs, contributing to more balanced power usage and mitigating performance bottlenecks due to single-motor saturation. Additionally, we designed a lightweight foot to enhance the robot's agility. We observed that applying the motor operating region as a constraint helps the policy network avoid infeasible areas during sampling. 
With the trained policy, KAIST Hound, a 45 kg quadruped robot, can run up to 6.5 m/s, which is the fastest speed among electric motor-based quadruped robots.
\end{abstract}

\section{Introduction}

Legged robots can leverage reinforcement learning (RL) techniques to enhance their ability to execute challenging tasks. The successful implementation of reinforcement learning algorithms has shown that robots can navigate through challenging environments, which include uneven, unstructured, and deformable terrains\cite{Science_JoonhoLee, Takahiro_Miki,RMA, SoftGround}.

These approaches are based on simulations where robots can collect data. By repeatedly interacting with their simulated environments, the agents can gradually improve their performance.

However, the simulation-to-reality (sim-to-real) gap remains one big challenge that needs to be addressed. This issue comes from the differences between the conditions in simulations and the unmodeled factors of real-world environments. Because of this gap, the real robot behavior and performance can change when transitioning from a simulation to a real world. To mitigate this gap, previous studies have employed domain randomization techniques. These techniques introduce a range of variability within simulation parameters, thereby enhancing the adaptability of reinforcement learning agents to the inherent unpredictability of real-world scenarios. There are attempts to randomize the training environment, such as terrain, so that the robot can walk robustly on various terrains \cite{Science_JoonhoLee, Takahiro_Miki}. To overcome the inaccurate system identification, such as torque constant or mass of the robot, noise is added to the robot parameters \cite{RMA, margolis2022rapid}. Hwangbo et al. \cite{Science_JeminHwangbo} mapped torque commands from a trained network to torques measured in the real world through a network to predict the behavior of actuators such as series elastic actuators, which is difficult to model in simulation. These difficulties include non-linearity of the actuator and lack of information from the manufacturer.

Besides the aforementioned research related to stable locomotion, there have been efforts to achieve highly dynamic motion through reinforcement learning. In challenging motions such as high-speed running, the motor output is used close to the boundary of its capability \cite{JinYongbin}. Therefore, studies have been conducted that reflect the characteristics of the motor in the simulator to find feasible trajectories in the real world for modeled acrobatic behaviors \cite{MIT_Humanoid}.

To integrate motor modeling into reinforcement learning simulators, we imposed set of linear inequality constraints representing feasible motor torque and angular velocity regions during training. This effectively clips the infeasible area in the real world to a feasible range for torque commands. These constraints correspond to the motor's operating range, readily available from its specification sheet.

By clipping motor output in the simulation for commands that would be infeasible, we prevented state transitions in the simulation that cannot be realized in the real world. This approach allows the robot to adapt torque commands that exceed the maximum capacity of the motor, even in the real world. We also observed that the proposed method helps the network avoid sampling the infeasible range of motor outputs during the learning process.

\begin{figure}[!t] 
  \centering
  \includegraphics[width=0.88\columnwidth]{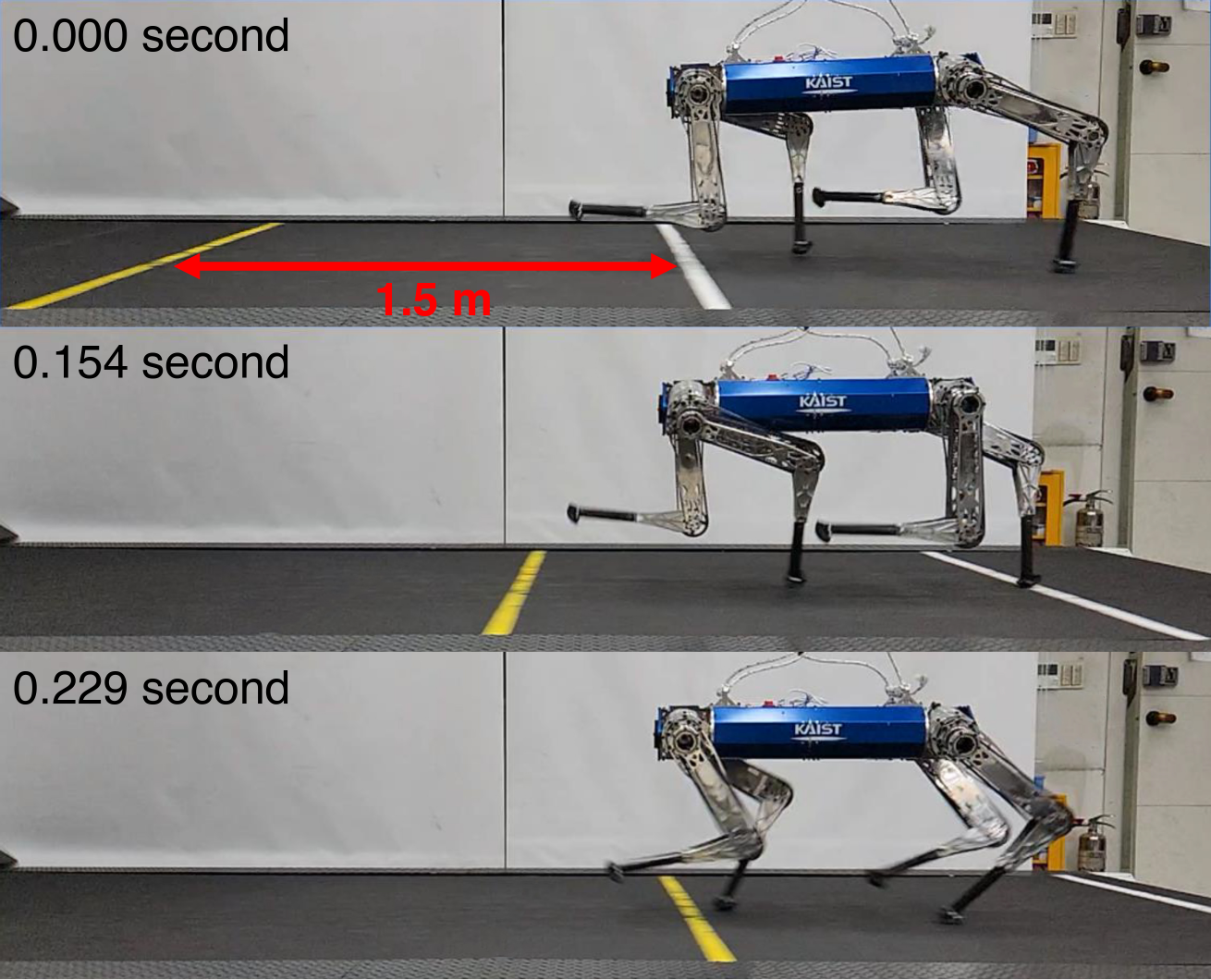}\\
\caption{KAIST Hound crosses a 1.5-meter distance between two lines on the treadmill in 0.229 seconds with an average speed 6.5 m/s. The snapshots are obtained from a video recorded with a rate of 240 frames per second.}
\label{fig:KAIST_Hound}
\end{figure} 

Our methodology was demonstrated using KAIST Hound \cite{HOUND}, which is a quadruped robot designed for high-speed running with optimized gear ratios. 
Using the proposed method, KAIST Hound, a 45 kg robot, can robustly run at speeds up to 6.5 m/s on a treadmill. This marks the fastest running speed recorded for an electric motor-based legged robot, surpassing the previous record of 6.4 m/s set by MIT Cheetah 2~\cite{MIT_Cheetah2}. Additionally, when controlled by reinforcement learning methodology, the highest speed achieved by a quadruped robot was previously reported to be 5 m/s~\cite{JinYongbin}.

The remainder of the paper is organized as follows. Section \ref{Methods} introduces the motor operating range constraints and overall training process. Section \ref{Results} verifies the proposed method with experimental results in both simulation and the real-world. Section \ref{Conclusion} concludes this paper.

\input{Methods.tex}

\input{Experiments.tex}

\section{Conclusion} \label{Conclusion}
This research demonstrates the importance of integrating the robot's physical design elements with reinforcement learning algorithms for high-speed locomotion. Our method incorporates actuator constraints in the training process. We evaluated the proposed method by conducting high-speed running with the designed and built in-house quadruped robot platform, KAIST Hound. The MOR constraints have successfully implemented a 6.5 m/s running test, where the robot accelerated from rest to 6.5 m/s in 8 seconds and remained on the treadmill for 35 seconds. By employing MOR constraints from specification sheets, our approach is adaptable to various tasks that require high power, such as jumping and carrying heavy payloads.

In future work, we can compare performance with methodologies that can explicitly incorporate inequality constraints into reinforcement learning \cite{kim2023rewards}.

\bibliographystyle{IEEEtran}
\bibliography{IEEEabrv, references}

\end{document}

%% file: format.tex
\usepackage[letterpaper, left=0.667in, right=0.667in,bottom=0.597in,top=0.792in]{geometry}
\usepackage{graphics} 
\usepackage{epsfig} 
\usepackage{mathptmx} 
\usepackage{times} 
\usepackage{amsmath} 
\usepackage{amssymb}  
\usepackage{mathtools}
\usepackage{booktabs}
\usepackage{threeparttable}
\usepackage{cite}
\usepackage[hidelinks]{hyperref}
\usepackage{algorithm}
\usepackage{algpseudocode}
\usepackage{graphicx}
\usepackage{epsfig}
\usepackage{multirow,makecell}
\usepackage{amsmath}
\usepackage{xcolor}
\usepackage{multirow}
\usepackage{array}
\usepackage{tabularx}

\usepackage{tikz}
\usetikzlibrary{bayesnet}

\usepackage{siunitx}
\usepackage{breqn}
\usepackage[colorinlistoftodos]{todonotes}

\usepackage[caption=false,font=footnotesize]{subfig}

\usepackage{lipsum}



%% file: Methods.tex
\begin{figure}[!t] 
  \centering
  \includegraphics[width=0.8\columnwidth]{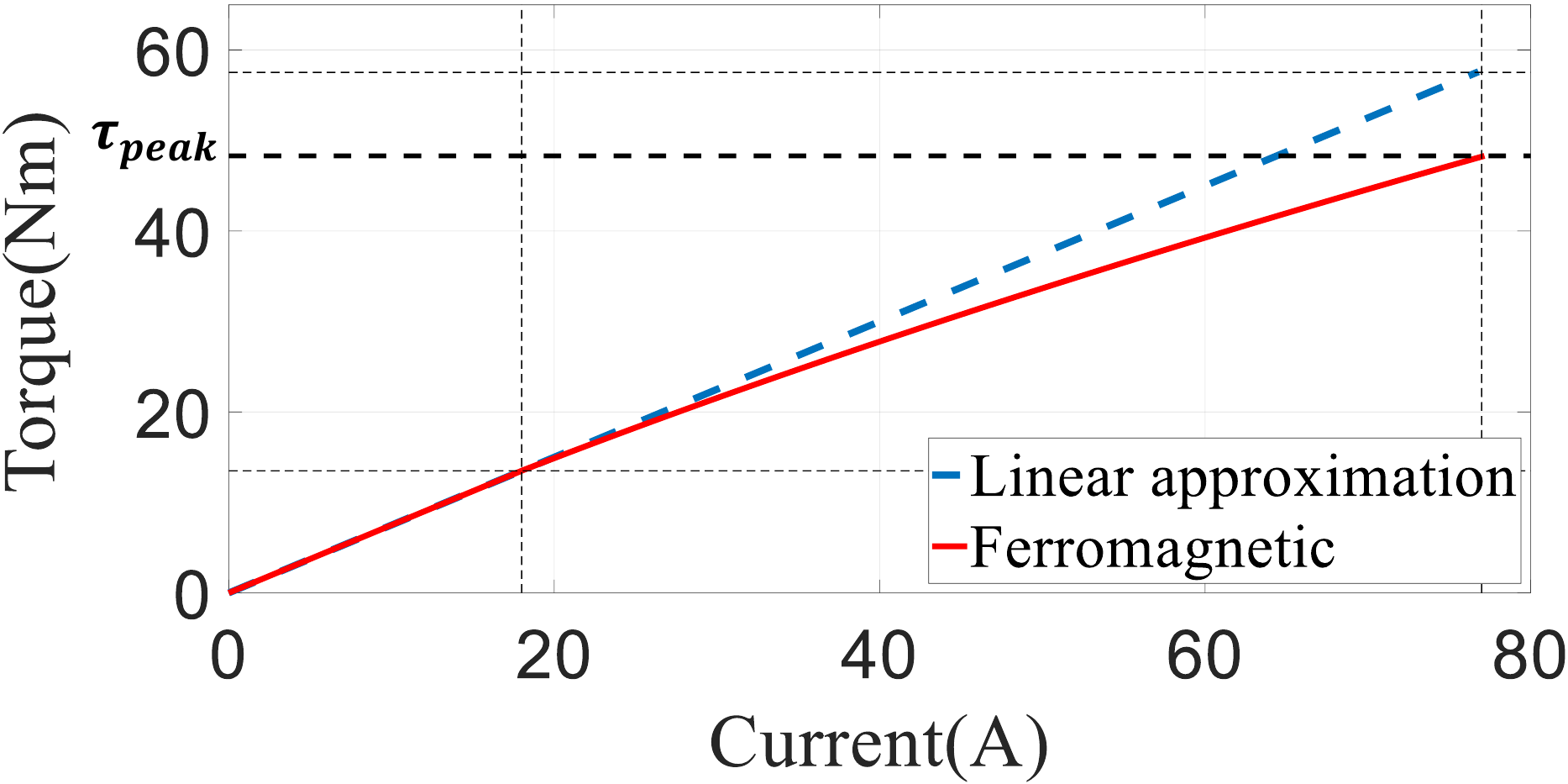}\\
\caption{Motor current versus motor torque of nonlinear ferromagnetic curve and dashed line for linear approximation.}
\label{fig:T-i}
\end{figure} 

\section{Methods} \label{Methods}

This section introduces our overall framework that enabled high-speed running of quadrupedal robots including policy training using RL and hardware modifications. Our RL framework employs linear inequality constraints to reflect more realistic motor operation regions and introduces a method for obtaining the gearbox matrix essential for the tranformation between motor space and joint space. The paper defines motor space as the input side without the reducer and joint space as the output side after the reducer. Further, we outline a gait reward strategy, and a learning framework focused on torque distribution and bridging the sim-to-real gap. Additionally, the development of a lightweight foot design aimed at enhancing robot agility and operational efficiency is discussed.

\subsection{Motor operating region (MOR)} \label{Motor_operating_region}

\begin{figure}[!t] 
  \centering
  \includegraphics[width=0.9\columnwidth]{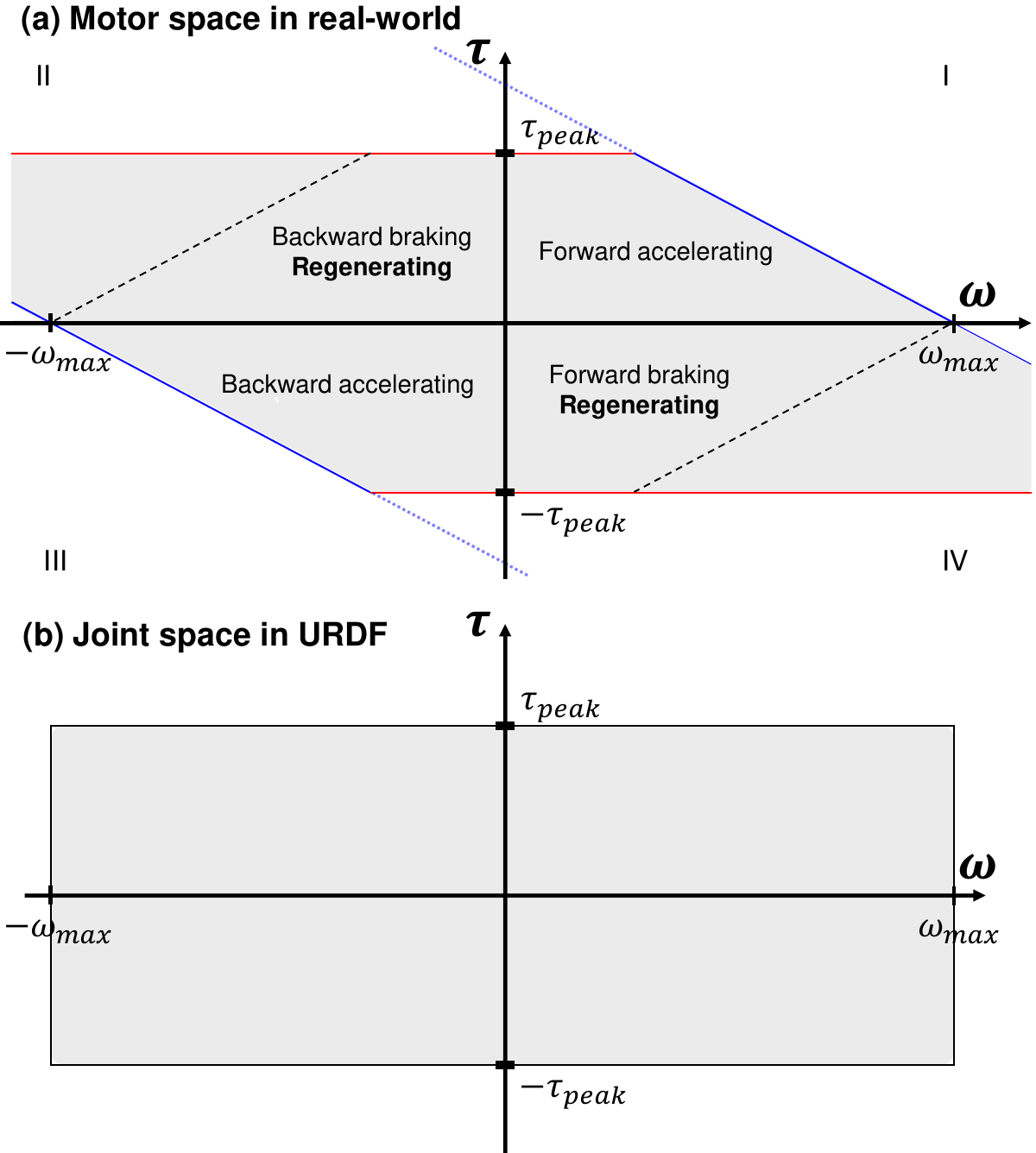}\\
\caption{(a) An operating region of BLDC motor with grey-colored based on the assumption that ignored the time derivative of the current. (b) Torque-velocity constraints defined by the URDF in joint space.}
\label{fig:MOR}
\end{figure}

The operating region of motors on $\tau - \omega$ space typically represents a trapezoid-like shape in the first quadrant as shown in Fig.~\ref{fig:MOR} (a), characterizing the motor’s performance \cite{MotorControl, chapman2012}. These geometries come from typical DC motor dynamics, considered the motor as an electrical circuit. For simplicity, assuming that the time rate of current change in a DC motor is negligible. This assumption is the same as when using a dynamometer to collect static data points and then plotting a $\tau - \omega$ curve \cite{Dynamometer}. Under quasi-stationary conditions, the equation of DC motor dynamics can be obtained as follows:

\begin{align}
\label{eq:MOR1}
V_{\text{applied}}  &= iR + L \frac{di}{dt} + V_{\text{BackEMF}}  \nonumber\\ 
                    &\approx iR + V_{\text{BackEMF}} \nonumber \\
                    &=  \frac{R}{K_t} \tau + \frac{1}{K_v} \omega,
\end{align}
where $V_{\text{applied}}$ is the voltage applied to counteract the motor's resistance $R$, $L$ is winding inductance, $i$ is the motor current, $V_{\text{BackEMF}}$ is the electromotive force produced by the motor due to its rotation, $K_v$ is the angular velocity constant, $K_t$ is the torque constant.

Equation \eqref{eq:MOR1} describes torque and angular velocity relationship of the motor under the applied voltage. Therefore, the operating region of the motor can be formulated as:

\begin{align}
\label{eq:MOR2}
& - V_{\text{bus}} \leq  \frac{R}{K_t} \tau + \frac{1}{K_v} \omega \leq V_{\text{bus}},
\end{align}
where $V_{\text{bus}}$ is the bus voltage. This relationship is represented by the area between the blue lines in Fig.~\ref{fig:MOR} (a).

For simplicity, some researchers assume that applied current and motor torque are linearly proportional \cite{Cheetah1_Hyun, Cheetah1_Seok}. However, in the case of a motor with a ferromagnetic core, which is widely used for efficiency and stability, the flux saturation of the core causes a nonlinearity, as shown in Fig.~\ref{fig:T-i} \cite{chapman2012}. The torque is limited to a peak torque $\tau_{\text{peak}}$ value, which is defined when the deviation from linearity is 20\% \cite{Robodrive}. The limit of the motor operating region by $\tau_{\text{peak}}$ is represented by the area between the red lines in Fig.~\ref{fig:MOR} (a). Note that $\tau_{\text{peak}}$ is not an absolute value but rather a recommended value that should not be exceeded to prevent the windings from burning out due to high temperature. When plotting this characteristic of a motor over the quadrants, the second and fourth quadrants are where the motor decelerates, and the sign of the work generated by the motor is negative. In physical interpretation, the motor acts as a generator, and the back EMF can contribute to the total voltage. In these quadrants(regenerative braking or generator mode), the operating range can be wider compared to the first and third quadrants, as the motor can potentially generate voltages through back EMF that exceed what the battery alone can provide. The resulting operating range of the motor is represented in Fig.~\ref{fig:MOR} (a).

In \cite{MIT_Humanoid}, where the $\tau - \omega$ curves for the motor modules were obtained through experiment, it enabled the precise calculation of actuator constraints for each time step in the simulation for model predictive control (MPC). However, this study did not consider regions besides the first quadrant of the torque-speed graph, which includes deceleration important for achieving dynamic motions.

\subsection{Transformation between joint space and motor space} \label{Transition between joint space and motor space}

\begin{figure}[!t] 
  \centering
  \includegraphics[width=1.0\columnwidth]{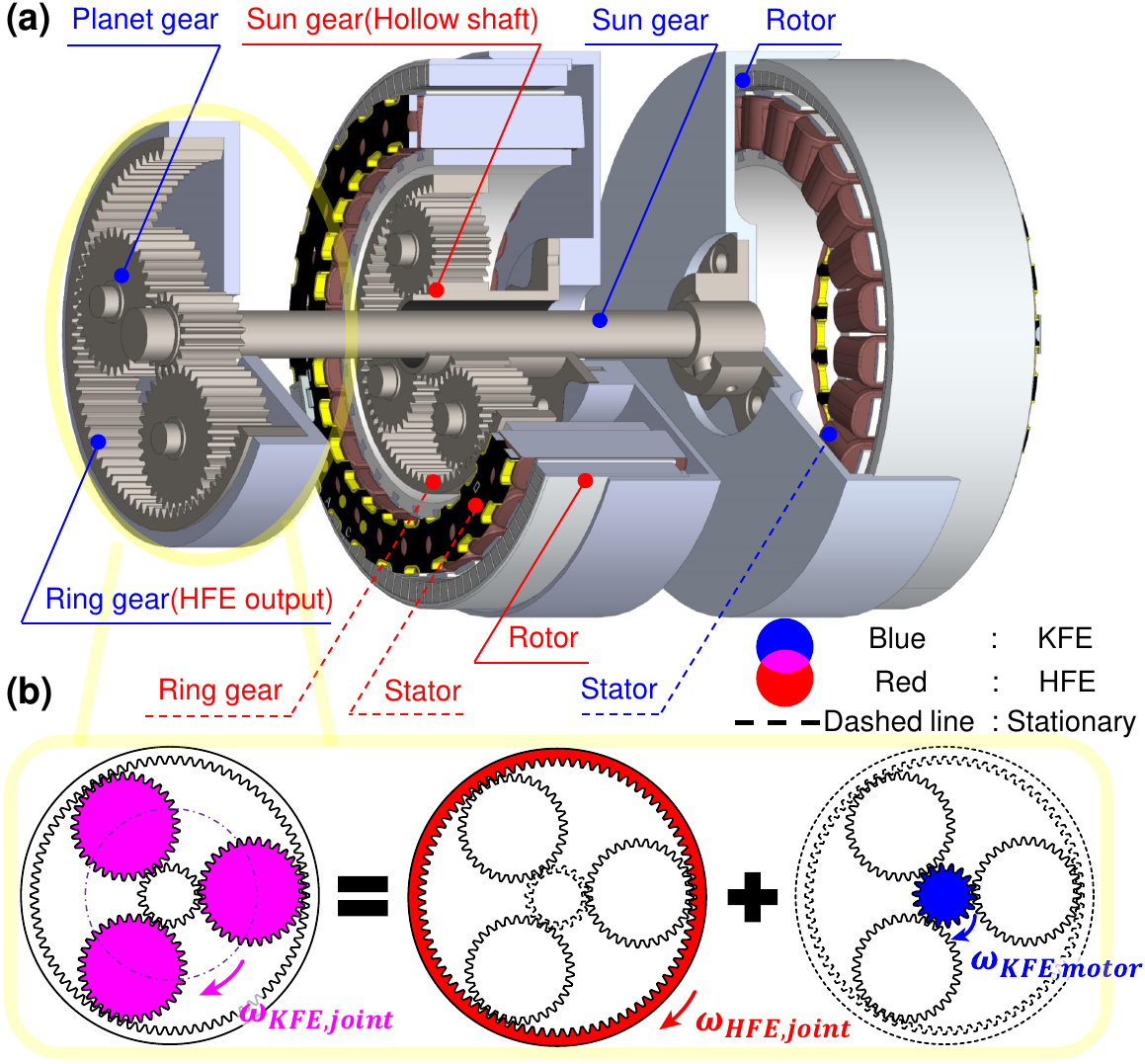}\\
\caption{(a) The conceptual design of KAIST Hound actuator. (b) Superposition of the angular velocity of the KFE and HFE.}
\label{fig:Gearbox}
\end{figure} 

URDF specifies the upper and lower bound of torque and velocity limits of each joint as constants, respectively. These geometries are rectangular, as shown in Fig.~\ref{fig:MOR} (b), and are different from Fig.~\ref{fig:MOR} (a). To reflect the system dynamics from the motor perspective, the parameters of the joint space should be converted to the motor space. URDF uses a joint angle configuration where all parts of the child link move with respect to the ground as much as the parent link moves with respect to the ground. This convention poses difficulty when it describes parallel configurations such as KAIST Hound \cite{HOUND}. As shown in Fig.~\ref{fig:Gearbox}, the hip flexion/extension(HFE) and knee flexion/extension(KFE) modules of KAIST Hound are composed of one piece. Since HFE is designed with a hollow shaft, when HFE rotates, there is no relative motion between the ground and input of the KFE.

The superposition principle was used to obtain the relationship between the input and output of the reducer as a serially connected joint angle configuration from a parallel actuator configuration. In the gearbox of the KFE, the sun gear is the input, the ring gear is stationary, and the planet gear is the output. However, the output part of the HFE is bonded to the ring gear of the KFE. Therefore, each rotation of the HFE input causes the ring gear of the KFE to rotate by $1/G_{KFE}$, where $G_{KFE}$ is the reduction ratio of the KFE joint. The rotation of the KFE ring gear by $1/G_{KFE}$, which is the stationary part of the KFE gearbox, denotes that the sun gear rotates by $-1/G_{KFE}$ from the perspective of the sun gear. Note that a negative sign indicates the reverse direction. The resulting motor velocity and motor torque obtained by the superposition principle are as follows:

\begin{equation}
\label{GearboxMatrix}
\begin{bmatrix}
\omega_{HFE, motor} \\
\omega_{KFE, motor}
\end{bmatrix}
=
\begin{bmatrix}
\frac{1}{G_{HFE}} & 0 \\
-\frac{1}{G_{HFE} G_{KFE}} & \frac{1}{G_{KFE}}
\end{bmatrix}^{-1}
\begin{bmatrix}
\omega_{HFE, joint} \\
\omega_{KFE, joint}
\end{bmatrix}
\end{equation}

\begin{align}
\label{GearboxMatrix2}
\begin{bmatrix}
\tau_{HFE, motor} \\
\tau_{KFE, motor}
\end{bmatrix} 
=
\begin{bmatrix} 
       \frac{1}{G_{HFE}} & 0 \\
       -\frac{1}{G_{HFE}G_{KFE}}  & \frac{1}{G_{KFE}} 
    \end{bmatrix}^{-T}
    \begin{bmatrix}
\tau_{HFE, joint} \\
\tau_{KFE, joint}
\end{bmatrix}
\end{align}

\subsection{Motor torque saturation} \label{Motor torque saturation}
In the training process, the desired joint position is output from the network and converted to joint torque by the PD controller with the desired joint velocity of zero. Due to the aforementioned limitation that the operating range of the real world motor is represented by a square shape in the joint space of the simulation, the converted joint torque might be infeasible for a real-world motor. Since the joint torque used in the simulation might not be feasible in the real-world motor, the behavior of the robot between the simulation and the real world can be changed. Therefore, we convert the joint torque command that violates the MOR into a motor torque and then clip the value according to equation \eqref{eq:MOR1} in the simulation. The saturated motor torque is converted to a saturated joint torque and applied to the robot.
One thing to note is that low-level commands correspond to currents in the real world and torques in simulation, respectively. To narrow this gap, the relationship between current and torque was considered. As explained in section.~\ref{Motor_operating_region}, motors with ferromagnetic cores on the real robot have a non-linear relationship between current and torque. Especially at high torque values, the deviation from linearity becomes larger, so we collected data sets of current and torque from single-axis motors and compensated for the current from quadratic approximation as plotted in Fig.~\ref{fig:T-i}.

\begin{figure*}
\centering
\includegraphics[width=2.0\columnwidth]{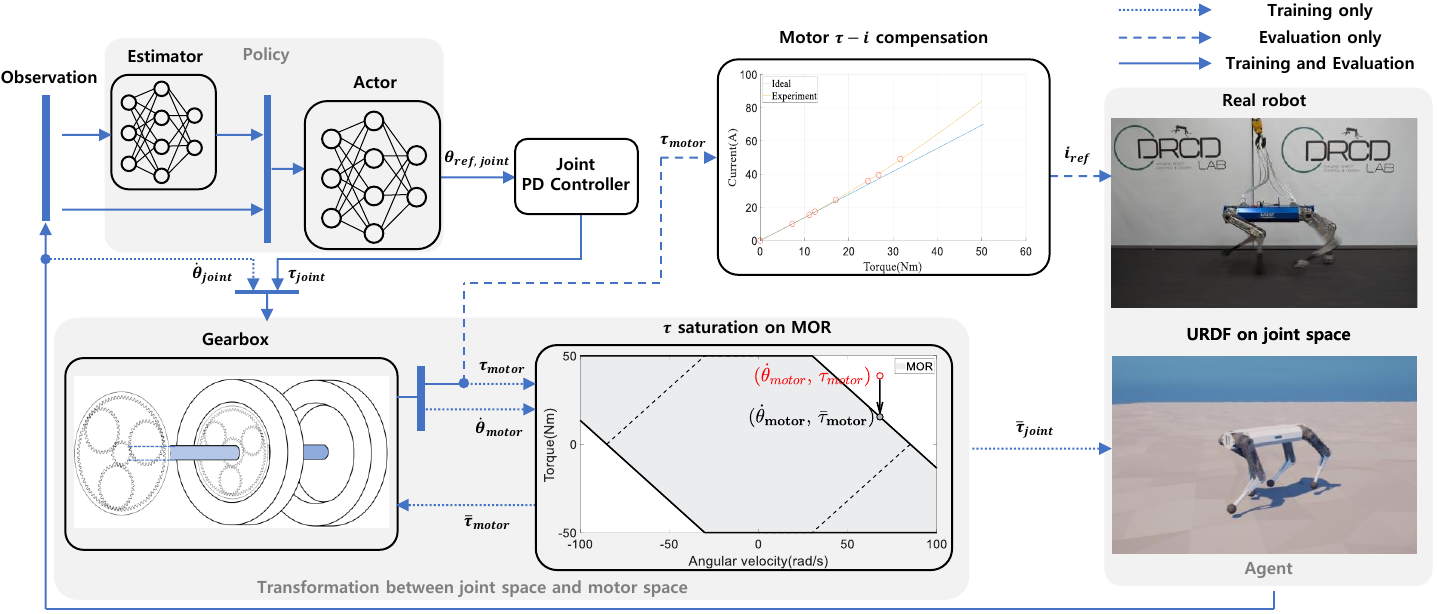}%
\label{fig:Framework}
\caption{The overall proposed reinforcement learning framework is shown. The action was expressed in terms of motor torque, and the torque was saturated by the MOR in the simulation. The nonlinearity of the current and torque due to the ferromagnetic core is compensated when the network performs inference.}

\label{fig:Framework}
\end{figure*} 

\subsection{Gait reward} \label{Gait reward}
The concurrent training of networks and the overall reward functions are based on \cite{Ji}. However, upon implementing this approach, we observed a phenomenon where motor power was biased towards one side, resulting in the robot's performance being limited by the saturation of one motor. To distribute the torque of the four legs evenly through a symmetrical gait, the gait reward is designed as shown in the TABLE \ref{tab:Reward}. The index f indicates the foot, $C_{f,i}$ represents the contact state of the foot for index i. When $C_{f,i}$ equals 1, the leg is in contact; when $C_{f,i}$ is 0, the leg is not in contact. When i is 1, 2, 3, and 4, it refers to the rear right (RR), rear left (RL), front right (FR), and front left (FL) legs, respectively. The effectiveness of the gait reward is demonstrated in the ablation study presented in the TABLE \ref{tab:Effectiveness}.

\subsection{Learning framework descriptions} \label{Learning framework descriptions}
\subsubsection{Overall structure} \label{Overall structure}
 Our framework is described in Fig.~\ref{fig:Framework}. The policy network and value network with layers of [256, 128, 64], and the estimator network consists of [256, 128] layers. As in \cite{Ji}, the observation, which is raw data from the sensor, inputs the estimator network. The body linear velocity, foot height, and contact probability from the estimator are merged with the raw data and input to the actor. The desired joint position from the actor is input to the joint PD controller with \( K_p = 50 \, \text{N} \cdot \text{m} \cdot \text{rad}^{-1} \) and \( K_d = 1.0 \, \text{N} \cdot \text{m} \cdot \text{s} \cdot \text{rad}^{-1} \) and converted to joint torque. The joint velocity and joint torque commands are converted through equations (\ref{GearboxMatrix})-(\ref{GearboxMatrix2}) and saturated to the boundary of MOR if the motor torque corresponding to the motor velocity violates the MOR. The saturated motor torque is converted to joint space by the gearbox and input to the robot. In the real world, the motor torque obtained from the network is converted to the input current of the motor by corresponding to the $\tau - $$i$ graph as shown in Fig.~\ref{fig:Framework}.

\subsubsection{Reward functions} \label{Reward functions}
 Our reward functions are defined as shown in the TABLE \ref{tab:Reward}. Bolded text indicates a new term added for high-speed running. Note that $k_{\text{contact}}$ is only 1 in the time step immediately before the contact occurs and 0 otherwise.

\begin{table}[h]
\centering
\caption{Reward Functions}
\label{tab:Reward}
\begin{tabular}{ll}
\hline
\textbf{Reward} & \textbf{Expression} \\
\hline
Linear velocity & \( C_V \exp(-||V^{\text{desired}}_{xy} - V_{xy}||^2) \) \\
Angular velocity & \( C_{\dot{\theta}} \exp(-1.5({\dot{\theta}}^{\text{desired}}_z - {\dot{\theta}}_z)^2) \) \\
\textbf{Contact velocity} & \( C_{\text{contact}} k_{\text{contact}} ||V_{f,z,i}||^2 \) \\
Foot slip & \(C_{\text{slip}} C_{f,i} ||V_{f,xy,i}||^2 \) \\
Foot clearance & \( C_{\text{clearance}} (\lnot C_{f,i}) (w p_{f,z,i} - w p^{\text{desired}}_{f,z})^2||V_{f,xy,i}||^{0.5} \) \\
Orientation & \( C_{\text{orientation}} (\text{angle}(\varphi_{\text{body},z}, \varphi_{\text{world},z}))^2 \) \\
Joint torque & \( C_\tau ||\tau||^2 \) \\
Joint position & \(C_q ||q_t - q_{\text{nominal}}||^2 \) \\
Joint speed & \(C_{\dot{q}} ||\dot{q}_t||^2 \) \\
Joint acceleration & \(C_{\ddot{q}} ||\dot{q}_t - \dot{q}_{t-1}||^2 \) \\
Action smoothness 1 & \(C_{\text{smoothness1}} ||q^{\text{desired}}_{t} - q^{\text{desired}}_{t-1}||^2 \) \\
Action smoothness 2 & \(C_{\text{smoothness2}} ||q^{\text{desired}}_{t} - 2q^{\text{desired}}_{t-1} + q^{\text{desired}}_{t-2}||^2 \) \\
Base motion & \(C_{\text{base}}(0.8V^2_z + 0.2|{\dot{\theta}}_x| + 0.2|{\dot{\theta}}_y|) \) \\
\textbf{Gait} & \(
\begin{cases} 
\text{if } C_{f,1} \land C_{f,4} \land \lnot C_{f,2} \land \lnot C_{f,3} \text{ : } C_{\text{gait}} \\
\text{else if } \lnot C_{f,1} \land \lnot C_{f,4} \land C_{f,2} \land C_{f,3} \text{ : } C_{\text{gait}} \\
\text{else : } 0
\end{cases}
\) \\
\hline
\end{tabular}
\begin{tabularx}{8.64cm}{|X|X|X|X|X|X|}
\hline
\multicolumn{6}{|l|}{\textbf{Reward Coefficients}} \\ \hline
\multicolumn{1}{|l|}{$C_V$} & \multicolumn{1}{l|}{6.0} & \multicolumn{1}{l|}{$ C_{\dot{\theta}}$} & \multicolumn{1}{l|}{3.0} & \multicolumn{1}{l|}{$C_{\text{gait}}$} & 1.2 \\ \hline
\multicolumn{1}{|l|}{$C_{\text{contact}}$} & \multicolumn{1}{l|}{-6.0}  & \multicolumn{1}{l|}{$C_{\text{slip}}$}  & \multicolumn{1}{l|}{-1.6e-1}  & \multicolumn{1}{l|}{$C_{\text{orientation}}$ }  & -3.0  \\ \hline
\multicolumn{1}{|l|}{ $C_{\tau}$ }  & \multicolumn{1}{l|}{-2e-4}  & \multicolumn{1}{l|}{$C_q$ }  & \multicolumn{1}{l|}{-0.75 }  & \multicolumn{1}{l|}{ $C_{\dot{q}}$ }  & -3e-4  \\ \hline
\multicolumn{1}{|l|}{$C_{\ddot{q}}$}  & \multicolumn{1}{l|}{-0.67e-2}  & \multicolumn{1}{l|}{$C_{\text{base}}$}  & \multicolumn{1}{l|}{-4}  & \multicolumn{1}{l|}{$C_{\text{smoothness1}}$}  & -2.5  \\ \hline
\multicolumn{1}{|l|}{$C_{\text{smoothness2}}$}  & \multicolumn{1}{l|}{-0.8}  & \multicolumn{4}{l|}{}\\ \hline
\end{tabularx}
\end{table}

We added a contact velocity reward that reduces the relative velocity between the ground and the foot just before the collision. It is designed to prevent unstable contact caused by the restitution coefficient between the ground and the foot, which is not observed in the simulation based on the hard contact model, RaiSim \cite{raisim}.

\subsubsection{Strategies for high-speed running in simulation} \label{Strategies for high-speed running in simulation}

The body mass, inertia, and center of mass positions, which are difficult to identify, were randomized. We also implemented a curriculum-based approach, progressively increasing the velocity command in the forward direction, which was uniformly sampled from \( \text{U}^1(-0.3, 1.5) \, \text{m/s} \) at the beginning. At the end of the learning process, this range was broadened to \( \text{U}^1(-1.4, 7.0) \, \text{m/s} \) as shown in Fig.~\ref{fig:NumberOfDataSamples}. At first, to allow the robot to react to abrupt decelerations and accelerations, we saved the robot’s final state from the previous episode and used it as the initial state for the next episode with a 25\% probability. We observed that when the gap between the robot’s current velocity and the command velocity was too large, the robot would go to a standstill to avoid terminating the episode while tracking. Therefore, when the episode starts from a 75\% stop, the command is increased gradually over the first second of the episode. For the remaining 25\%, we added an additional condition to ensure that the command velocity difference between the previous episode and the current episode does not exceed 2 m/s. As the robot runs at higher speeds, it receives larger penalties due to a reward term with a negative coefficient, such as joint torque and joint speed. Therefore, as the robot reaches higher speeds, it increasingly tends to abandon command tracking. To encourage the robot to follow high-speed commands, the weights of linear velocity and angular velocity in the reward term were gradually increased.

\subsection{Lightweight foot design} \label{Lightweight foot design}

\begin{figure}[!t] 
  \centering
  \includegraphics[width=0.6\columnwidth]{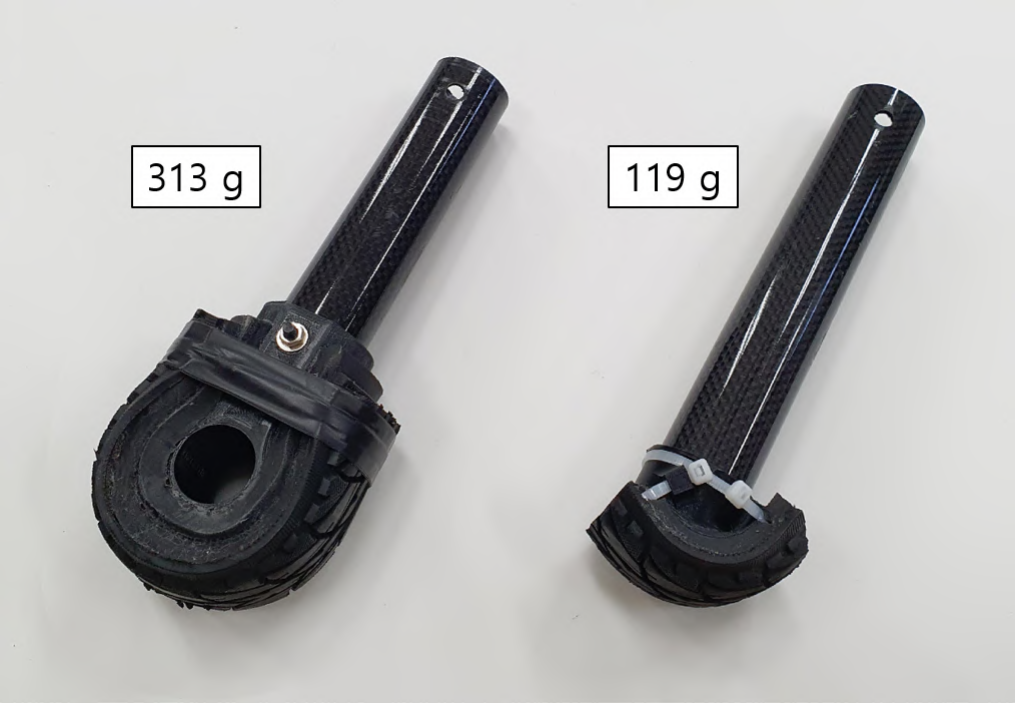}\\
\caption{Customized foot for high-speed running. Left : The original design in a cylindrical shape. Right : A lightweight design in a part of a spherical shape.}
\label{fig:FOOT}
\end{figure} 

We designed a specialized foot for high-speed running with reduced leg inertia, which can decrease impact forces and increase mechanical bandwidth  \cite{Cheetah_1, Cheetah2_Hardware}. Since the foot is located at the end of the leg link, a lightweight design is important to reduce the moment of inertia of the knee joint. Most quadruped robots are designed with feet that are either spherical or cylindrical in shape \cite{HOUND, BigDog, Anymal, Cheetah3,Katz_2019_minicheetah, Spot, Aliengo,hutter2012starleth}, which allow for a constant rolling contact that helps simplify contact dynamics to a point or line. However, as the foot approaches a perfect sphere or cylinder shape, the weight also increases. In the case of high-speed running, the robot only needs a foot shape that corresponds to the attack angle and detach angle between the flat ground and the foot. Therefore, we investigated the running data in simulation and designed the foot with a minimal sphere shape that ensures contact with the ground. As a result, the mass of the calf was reduced to 38.0\%, and the inertia in the pitch direction was reduced to 37.4\% of its original.

%% file: Experiments.tex
\section{Results} \label{Results}

\subsection{Evaluation in simulation} \label{Evaluation in simulation}

We investigated the effectiveness of MOR constraints for robot performance in simulation. First, we trained a controller without MOR constraints as a baseline. The robot reached the target speed from rest with an acceleration of 1 m/$s^2$. The motor operating region is violated when running at high speed in the simulation, as shown by the red line in Fig.~\ref{fig:Big_figure} (c). To reflect the behavior of actual motors in the simulation, we clipped the torque commands that violate the MOR, within the operational limits.

\begin{figure}[!t] 
  \centering
  \includegraphics[width=0.9\columnwidth]{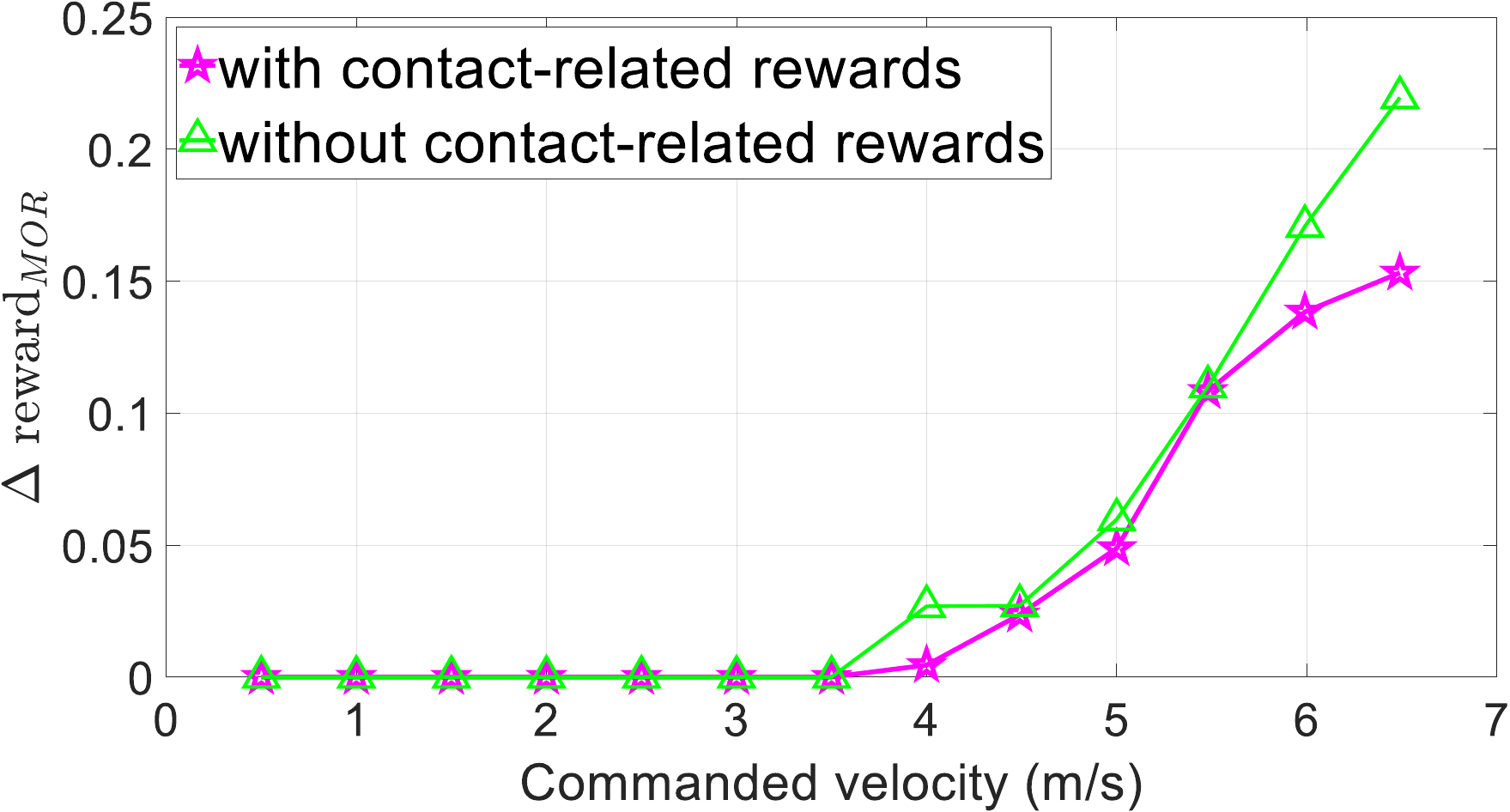}\\
\caption{The graph shows that when the policy learned without a MOR constraint is evaluated with a MOR constraint in the simulation, the performance drop occurs from 4 m/s using torque that violates the MOR. The two lines exhibit a similar trend in $\Delta$ reward$_{MOR}$, even when excluding contact-related rewards that are challenging to estimate in the real world.}
\label{fig:simulation_reward_compare}
\end{figure}

We used average reward as our performance metric, which indicates how close the robot's locomotion is to our designed behavior. The average reward was computed for 3 seconds after the robot speed converged. The result is shown in Fig.~\ref{fig:simulation_reward_compare}. Note that $\Delta$ reward$_{MOR}$ = (Policy evaluation without MOR) - (Policy evaluation with MOR). As the speed increases, the average reward, including command tracking performance, drops under the MOR constraint. Since the velocity below 3 m/s was achieved with torque commands inside the MOR, there was no performance drop.

In the real world, contact probabilities from a state estimation network have been observed to be noisy at high speeds. Therefore, rewards that require contact information are difficult to use as a metric for robot performance. To address the aforementioned problem, we also investigated the difference of average reward without contact-related rewards. It showed a trend where the gap becomes larger as the speed increases, similar to the difference in average reward. These results show when the trained network without MOR constraint is evaluated in the real world, the torque output limitation can cause a performance drop.

\begin{figure*}
\centering
\includegraphics[width=2.0\columnwidth]{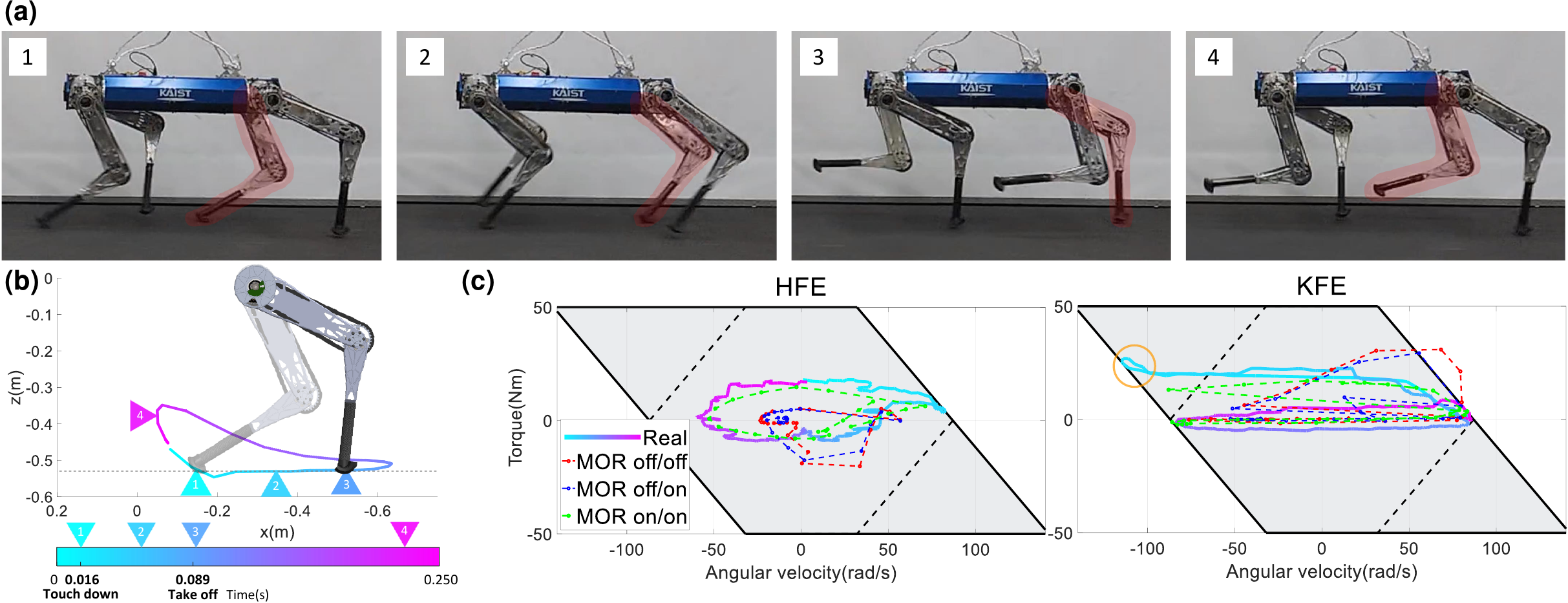}%
\label{fig:Big_figure}
\caption{ These figures provide detailed data of the rear right leg collected while running at 6.5 m/s in one period. (a): The red shading in the four snapshots represents the movement of the rear right leg. (b): Foot trajectory relative to the shoulder joint. The one cycle is 0.25 seconds, and the color gradually changes from cyan to magenta to indicate the position of the trajectory over time. (c): The $\tau - \omega$ trajectories for 6.5m/s running in simulation and real-world. MOR off/off: Training without MOR constraints and evaluation without MOR constraints. MOR off/on: Training without MOR constraints and evaluation with MOR constraints. MOR on/on: Training with MOR constraints and evaluation with MOR constraints.}
\label{fig:Big_figure}
\end{figure*}

\subsection{System setup} \label{System setup}

We used an AMD Ryzen Threadripper PRO 5995WX and a NVIDIA GeForce RTX 3080 Ti for the training process. The control policy requires about 6 hours to complete its training and operates at a frequency of 100 Hz. The on-board PC for control policy and estimation algorithms operates on a ThinkStation P350 Tiny. State estimation for body orientation is conducted by \cite{Mahony}. User inputs are transmitted to the high-level PC in real time via a Taranis X9D Plus transmitter with a X8R receiver. A commercial motor controller, Elmo Platinum Twitter 100 V/70 A, communicates with the high-level PC at a rate of 2 kHz through EtherCAT. The robot is powered by two different types of Lithium-Ion battery packs. For driving the motor, a 21700 Li-ion battery pack configured in a 20S 2P, 20 batteries in series, and two sets of these in parallel, is employed, offering high power within the limited body space. The other electrical parts are powered by a separate 18650 Li-ion battery pack in a 6S 3P configuration. The motor battery was charged to 81 V with a margin from the maximum charge of 84 V to permit reverse current from the motor.

\subsection{Experimental results} \label{Experimental results}

\subsubsection{Indoor experiments} \label{Indoor experiments}

\begin{table}[h]
\centering
\caption{Ablation study for the real-robot}
\label{tab:Effectiveness}
{
\footnotesize
\setlength{\tabcolsep}{0pt}
\begin{tabular}{
  | >{\centering\arraybackslash}m{1.5cm}
  | >{\centering\arraybackslash}m{1.5cm} 
  | >{\centering\arraybackslash}m{2cm} 
  | >{\centering\arraybackslash}m{2cm}   
  | >{\centering\arraybackslash}m{1cm}|} 
\hline
& \shortstack{without \\ gait reward} & \shortstack{without \\ customized foot} & \shortstack{without \\ MOR constraint} & \shortstack{\textbf{Ours}} \\ \hline
Best record (m/s) & 6.0 & 5.5 & 4.5 & \textbf{6.5} \\ \hline
\end{tabular}
}
\end{table}

We investigated how three major factors affect the performance of high-speed running.

\begin{itemize}
    \item Without gait reward: To verify the performance drop caused by the saturation of the power of single-motor, we set the coefficient of gait reward to zero. The gait reward is designed to induce an even distribution of the motor's power through a symmetrical gait.
    \item Without customized foot: To verify whether the reduced inertia due to the lightweight design of the foot is effective for high-speed running, we train the robot model with an original foot, which is cylindrical shape. The learned policy is used to control a robot with an original foot to measure its speed.
    \item Without MOR constraint: To validate the effect of the MOR constraint, we exclude clipping for torque commands that violate the MOR in the training process.
\end{itemize}

The robot ran at 0.5 m/s intervals on a treadmill with the trained controllers. We considered the trial at each speed interval a success if the robot maintained that speed  on the treadmill for 10 seconds. As shown in table~\ref{tab:Effectiveness}, the controller trained with our proposed method achieved the highest speed of 6.5 m/s. When we increased the treadmill speed to 7 m/s, the robot was pushed backward, and its feet went off the treadmill, so 7 m/s was considered a failure. The other three controllers recorded the speed of 6.0, 5.5, and 4.5 m/s respectively which are slower than  6.5 m/s. 

\begin{figure}[!t] 
  \centering
  \includegraphics[width=0.9\columnwidth]{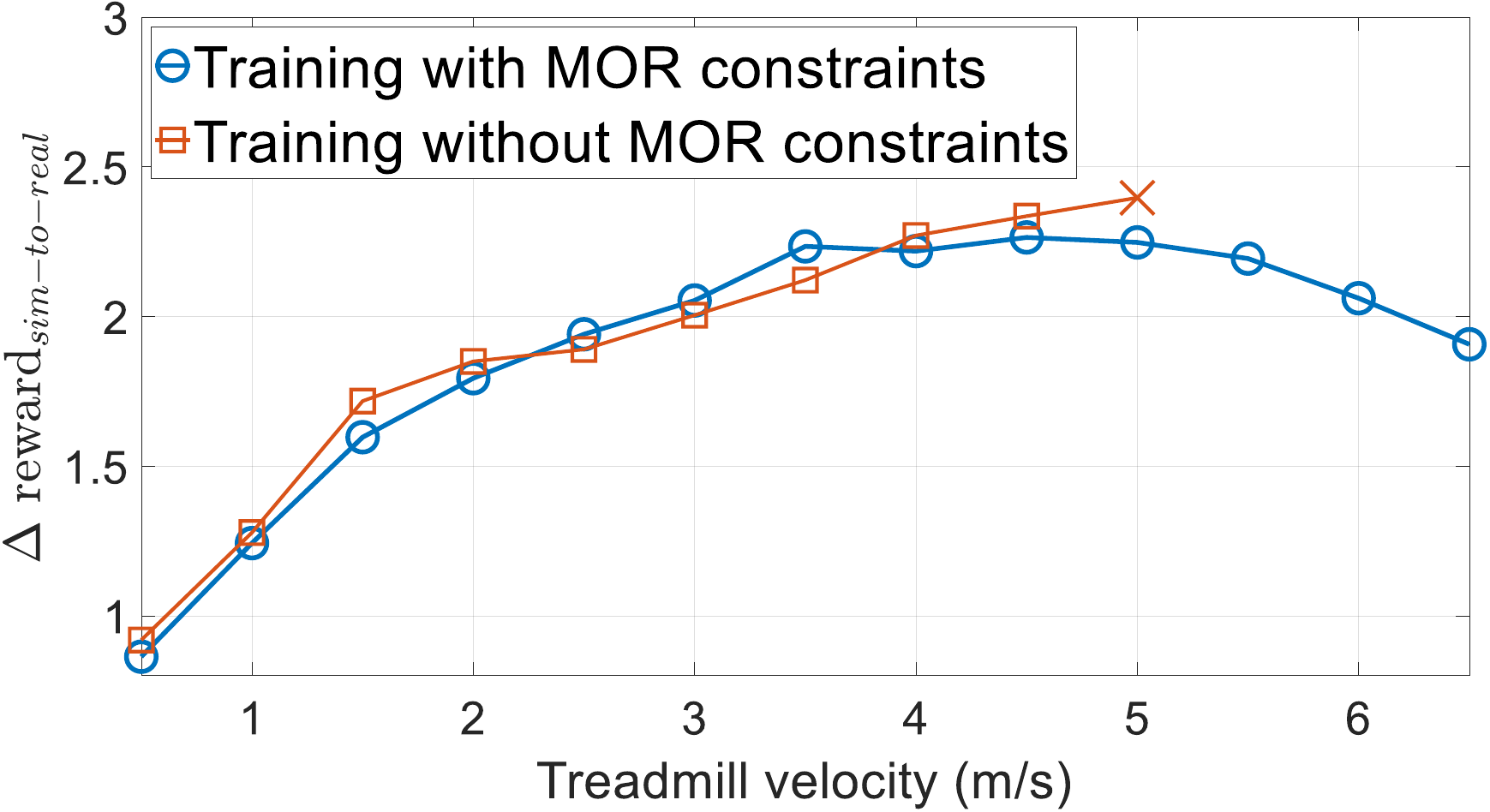}\\
\caption{The graph shows that applying the MOR constraint reduces the average reward when evaluating a policy learned without the MOR constraint in simulation. The controller trained without MOR constraints fell while the robot was running at 5 m/s.}
\label{fig:real_reward_compare}
\end{figure}

To further demonstrate the effectiveness of the MOR constraint, we computed the reward gap between the simulation and the real world from two policies. The two policies were trained with the MOR constraint and without the MOR constraint, respectively. Note that both policies can achieve 6.5 m/s in simulation. $\Delta$reward$_{sim-to-real}$ = (policy evaluation in simulation) - (policy evaluation in the real world) is defined as a performance metric, excluding the term computed by contact information, which is noisy at high speeds.  Since the robot runs on a treadmill, it is difficult to measure linear velocity with a motion capture system. Therefore, we use the linear velocity output from the estimator network trained by supervised learning with the true value in simulation and representing low estimation error to calculate the reward \cite{Ji}.

The results are shown in Fig.~\ref{fig:real_reward_compare}.  In the real world, as the robot runs faster, the higher $\Delta$reward$_{sim-to-real}$ due to the increase in unmodeled dynamics \cite{from_minicheetah_highspeed_1, from_minicheetah_highspeed_2} that are not included in the simulation. As shown in Fig.~\ref{fig:simulation_reward_compare}, the  $\Delta$reward$_{MOR}$ of the two policies tends to differ at high speeds above 3.5 m/s, where the motor uses torque that violates the MOR.

For the policy trained with MOR constraints, $\Delta$reward$_{sim-to-real}$ does not increase after 3.5 m/s. However, the robot controlled by the learned policy without MOR constraints continued to increase in $\Delta$reward$_{sim-to-real}$ and fell at 5 m/s. This result shows that enforcement of actuator limits helps the simulation behavior resemble the real world when the motor uses high power.

\subsubsection{Analysis at the top speed} \label{Analysis at the top speed}
At a running speed of 6.5 m/s, the cost of transport (CoT) for 15 strides, calculated using mechanical power and joule heating, is 0.29. Fig.~\ref{fig:Big_figure} (a) shows snapshots of the four distinct behaviors during one cycle at top speed. The foot trajectory of the rear right leg over time is plotted in Fig.~\ref{fig:Big_figure} (b) as a change in color, and the $\tau - \omega$ trajectories of the motor are plotted in Fig.~\ref{fig:Big_figure} (c). The dashed line shows the values from the simulation, and the solid line shows the real-world experimental results. The measured velocity and torque of the motor in the experiment are comparable to those in the simulation. The circled region in orange in the KFE in Fig.~\ref{fig:Big_figure} (c) corresponds to the early stage of the stance phase, where negative energy regeneration occurs. It is observed that the  $\tau - \omega$ trajectory at the region satisfies the MOR constraints by following the slope of the boundary of the MOR constraints in the second quadrant. This result shows that the motor operation well respects MOR constraints which is wider in the 2nd quadrants compared to the 1st quadrants, as described in section \ref{Motor_operating_region}.

\subsubsection{Outdoor experiments} \label{Outdoor experiments}

\begin{figure}[!t] 
  \centering
  \includegraphics[width=1.0\columnwidth]{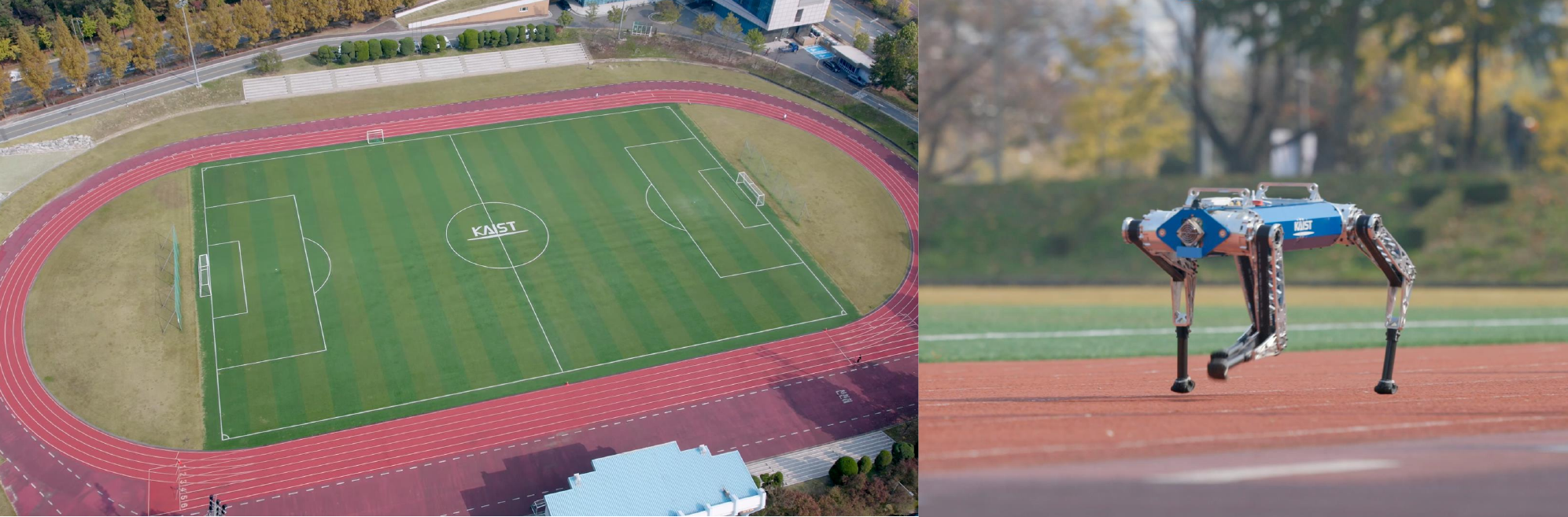}\\
\caption{The presented controller was tested on official athletic track.}
\label{fig:Guinness}
\end{figure} 

We also demonstrated the robot's performance on an official athletics track composed of polyurethane, as shown in Fig.~\ref{fig:Guinness}. The robot started running with an initial command of 1.2 m/s from rest and accelerated to 5.9 m/s, based on the estimator network, over 6.7 seconds. It took 19.87 seconds to run 100 meters, returning to the same pose having crossed the finish line. The robot is operated with the same command as when it achieved 6.5 m/s on the treadmill but at a slightly lower speed. We suppose that the softness of the urethane track contributes to the drop in maximum speed \cite{SoftGround}. A total of nine attempts were made, and the robot completed the 100-meter run without falling over all nine times.

\subsection{Discussion} \label{Discussion}

\begin{figure}[!t] 
  \centering
  \includegraphics[width=1.0\columnwidth]{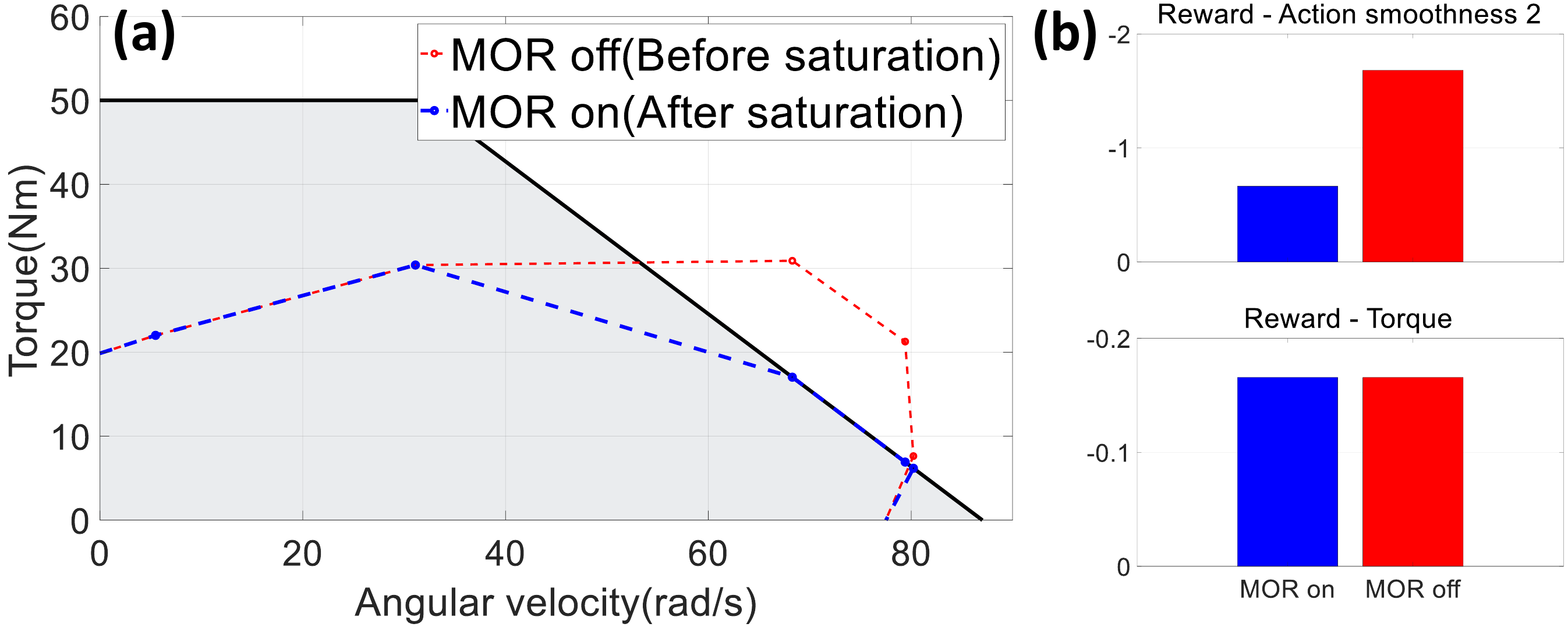}\\
\caption{(a) The red and blue dotted lines that are the $\tau - \omega$ output of the KFE motor when the robot runs in the simulation at 6.5 m/s subject to the MOR constraint. The red line is the trajectory of the torque command from the network output, and the blue line is the trajectory of the resulting saturated torque command by the MOR constraints. (b) The blue dotted line is more penalized by the action smoothness reward, but the torque applied to the robot is the same as the red trajectory.}
\label{fig:MOR_reward}
\end{figure} 

\begin{figure}[!t] 
  \centering
  \includegraphics[width=1.0\columnwidth]{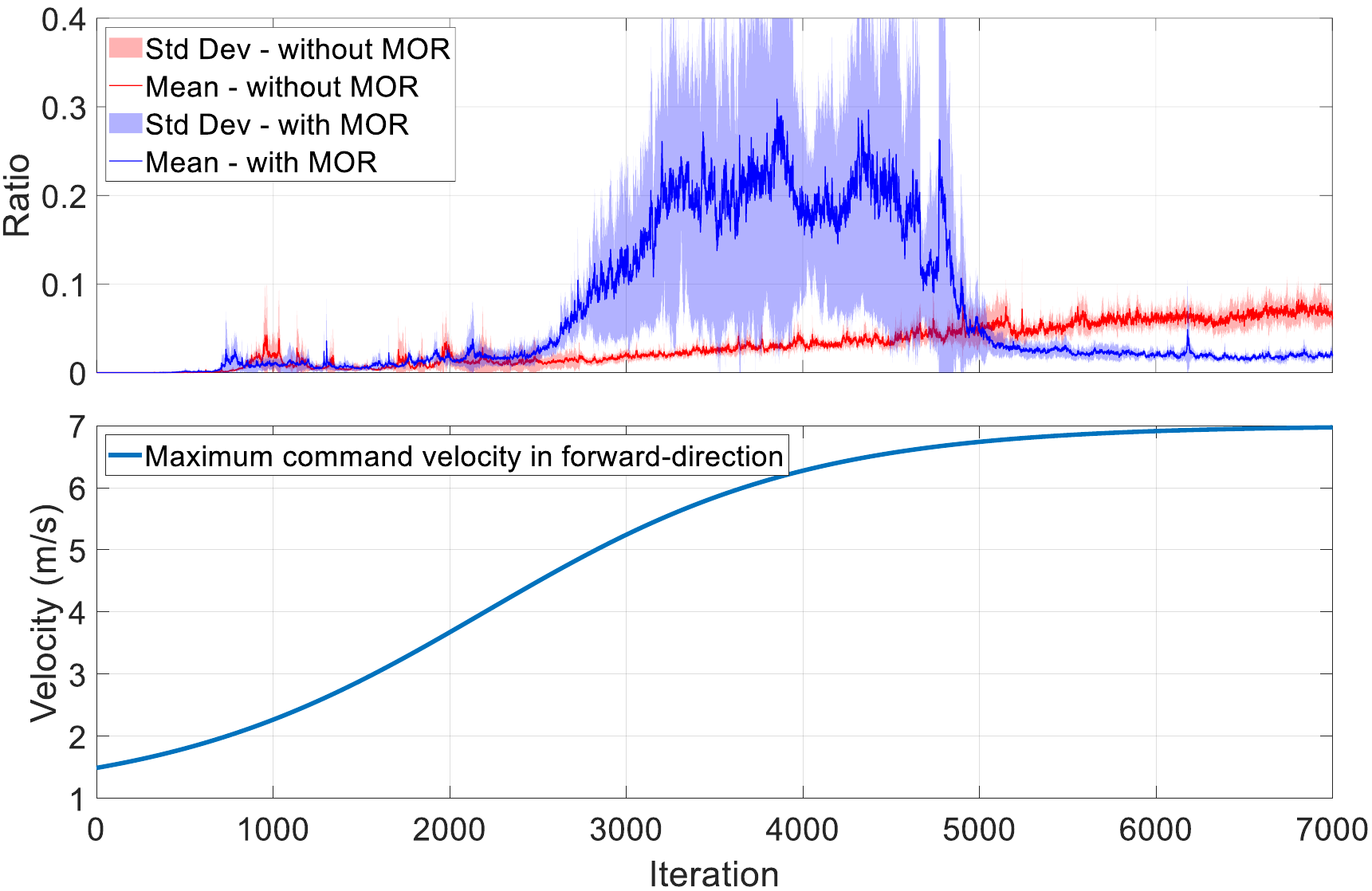}\\
\caption{Upper: Learning curves of different algorithms. The curves show the ratio of the number of torque commands sampled at each iteration that violated the motor operating range. The results shown are obtained from four different random seeds. Lower: Maximum value of the linear velocity command range.}
\label{fig:NumberOfDataSamples}
\end{figure} 

The two colored $\tau - \omega$ trajectories are shown in Fig.~\ref{fig:MOR_reward}. The red line shows the KFE motor output when the robot runs at a speed of 6.5 m/s by the same policy network which is trained without MOR constraints as the red line in Fig.~\ref{fig:Big_figure}. The red trajectory includes torque commands that violate the MOR. The blue line indicates that the torque command in the red line is saturated by the MOR constraint.

The output of the policy network is the desired joint positions, which are converted to joint torques by the PD controller, and the corresponding joint torques are converted to motor torques by the gearbox and applied to the robot. Since the output of the policy network that violates the MOR is clipped to the feasible region by the MOR constraint, the negative reward of the motor torque is the same for both trajectories. However, a sudden action change to produce a torque command that exceeds the MOR is highly penalized by the action smoothness reward. Therefore, the MOR constraint guides the policy network to avoid infeasible torque commands.

It is also observed by the number of data samples that violate the MOR during the learning process shown in Fig.~\ref{fig:NumberOfDataSamples}. In 400 parallel environments, each episode is 4 seconds long, and the control is done every 0.01 seconds, resulting in 160,000 data samples in every iteration. The vertical axis in Fig.~\ref{fig:NumberOfDataSamples} shows the ratio of torque commands exceeding the MOR out of 160,000 data samples. In the early stages of learning low-speed commands that require less motor power, the number of torque commands that violate the MOR is nearly zero, but as the commands increase, the number of data samples that violate the MOR increases. However, the ratio of data samples that violate the MOR in the presence of the MOR constraint converges to a relatively low value compared to learning without the MOR constraint. Therefore, the MOR constraints in the training process help the network to prevent infeasible output commands to the motor. For our study, the robot controller, which was trained without action smoothness reward, demonstrated vibrations when operated on the actual robot.